# A Multi Hidden Recurrent Neural Network with a Modified Grey Wolf Optimizer

[Tarik A. Rashid[1, 2*]; Dosti K. Abbas[3]; Yalin K. Turel[4]]

[[1]Computer Science and Engineering Department, University of Kurdistan Hewler, Kurdistan, Iraq,

[2]Software and Informatics Engineering, Salahaddin University-Erbil, Kurdistan, Iraq;

[3]Faculty of Engineering, Soran University, Kurdistan, Iraq;

[4]Department of Computer Education and Instructional Technology, Firat University, Elazig, Turkey]

*Corresponding author: tarik.ahmed@ukh.edu.krd

## Abstract

Identifying university students' weaknesses results in better learning and can function as an early warning system to enable students to improve. However, the satisfaction level of existing systems is not promising. New and dynamic hybrid systems are needed to imitate this mechanism. A hybrid system (a modified Recurrent Neural Network with an adapted Grey Wolf Optimizer) is used to forecast students' outcomes. This proposed system would improve instruction by the faculty and enhance the students' learning experiences. The results show that a modified recurrent neural network with an adapted Grey Wolf Optimizer has the best accuracy when compared with other models.

## Introduction

In education management, student performance prediction and classification systems are important tools. They warn students who did not perform well or those with at risk performance and assist students in averting and overcoming most of the problems they face in meeting their objectives. Yet, there are challenges in gauging students' performance, since academic performance depends on various elements, such as demographics, personalities, education background, psychological issues, academic progress and other environmental variables [1].

Statistical methods, data mining, and machine learning techniques are used for extracting useful information related to educational data. This is known as 'educational data mining' (EDM) [2]. EDM uses academic databases and constructs several techniques for identifying unique patterns [3, 4] to benefit academic planners in educational institutions by identifying ways to improve the process of decision-making.

Academic performance research studies mostly have been carried out using classification and prediction methods. The task of classification is regarded as a process of determining a model in which data are classified into categories [5]. Neural networks are part of machine learning and are regarded as one the best means of modeling classification problems that imitate human neural activity. The basic concept of neural networks was first proposed in 1943 [6]. İt is worth mentioning that various classes of neural networks have been developed, such as feed-forward networks [7], radial basis function (RBF) networks [8], Kohonen self-organizing networks [9], spiking neural networks [10], and recurrent neural networks [11].

Neural networks are trained with back-propagation learning algorithms, which are usually slow and thus need higher learning rates and momentum to achieve faster convergence. These approaches perform well only if the incremental training is required. However, they are still too slow for 'real life' applications. Nonetheless, the Levengerg-Marquardt model is still used for small- and medium-size networks. The lack of available memory is what prevents the use of faster algorithms. Back-propagation is a deterministic algorithm that tackles linear and non-linear problems. Yet, back-propagation and its variations may not always find a solution. Another problem associated with back-propagation algorithms is selecting an appropriate learning rate, which is a complicated issue. For a linear network, a too-fast learning rate would cause unstable learning; on the other hand, a too-slow learning rate causes an excessively long training time. The problem is more complex for nonlinear multilayer networks, as it is difficult to find an easy method for selecting a learning rate. The error surface for nonlinear networks is also more challenging than that of linear networks [12]. On the other hand, using neural networks with nonlinear transfer functions would present several local minimum solutions in the error surface. Thus, it is possible for a solution in a network to become 'stuck' in a local solution. This can occur depending on the initial starting conditions. It is worth mentioning that having a solution in the local minima might be a satisfactory solution if the solution is close to the global minimum. Otherwise, the solution is incorrect. In addition, the back-propagation learning algorithm does not produce perfect weight connections for the optimal solution. In this case, the network needs to be reinitialized repeatedly to guarantee that the best solution is obtained [13, 14].

In contrast, there are nature-inspired algorithms, which are derived from the natural behavior of animals. These algorithms are stochastic. The essential element that is imported into these algorithms is randomness. This means that the algorithms use initial randomized solutions that are then improved through a sequence of iterations that avoid high local optima. Further, a multilayer neural network is subtle when it comes to deciding on selecting hidden neurons. There is an under-fitting problem that may arise when a small number of hidden neurons are used; also, overfitting can arise when too many hidden neurons are used. An alternative to a multilayer neural network is a recurrent neural network (RNN). An RNN uses fewer hidden neurons because it has a context layer for preserving previous hidden neuron nets. Therefore, the network is more stable and can successfully handle temporal patterns [15].

Recurrent neural networks can imitate the human brain to forecast student performance while considering the students' social and academic histories. This work presents a modified recurrent neural network and a modified Grey Wolf Optimizer. The latter is used for optimizing a modified former. The research work is structured as follows. Related works are described in section two. The preliminaries of the study are introduced in section three. The proposed method is described in section four. In section five, the results and discussion are presented. Finally, section six presents the conclusions of the work.

## Related works

In this section, the related works of two concepts are discussed in two parts, as follows: the state of the art applications for forecasting student performance and the state of the art grey wolf optimizer applications with/without neural networks.

# The state of the art applications for forecasting student performance

A neural network model was used for forecasting student performance in terms of Cumulative Grade Point Average (CGPA). The researchers used a dataset that contained the records of 120 students registered at Bangabandhu Sheikh Mujibur Rahman Science and Technology University. A neural network was trained with the Backpropagation Levenberg Marquardt learning algorithm. The network was trained with a dataset allocated for training, validating and testing sets for reducing the percentage of error. They concluded that the early performance of students depends on academic and outside influences, for example, social media, living area conditions, communication, etc. [13]. It was reported that neural networks have been successfully used for forecasting student performance better than the decision table, decision tree, and linear regression. The ID3 classification method was used for forecasting student performance. The task for extracting information related to student performance was conducted at the end of the examination. This study used data collected from VBS Purvanchal University. Significant elements of the information, such as the class test, attendance, assignment marks, and seminar type, were collected [14].

Significant attributes such as the study environment and social demographics that influence dropout rates at the Open Polytechnic of New Zealand were explored in [16]. The study environment includes the course program and course block. The social demographics included features such as gender, age, disability, ethnicity, education, and work status. The dataset included 450 patterns, and the data were obtained for a course in the period of 2006-2009. The main task was to perform a quality analysis of the results of the study. The most relevant features for student success and failure were identified based on data mining approaches such as feature selection and classification trees. The research produced the following results: It was found that the course program, ethnicity and course block were the most relevant features in distinguishing effective students from non-effective ones. A CART (classification and regression tree) produced better results than the other classification tree growing methods. It was also concluded that the gain diagram and cross validation generated approximated risk, which indicated that all trees are not appropriate.

The study presented in [17] was related to the lecturers' performance. A dynamic and smart system, using both multiple and single soft computing classifier techniques, was utilized for forecasting the lecturers' performance at the College of Engineering, Salahaddin University-Erbil. The collected dataset consisted of continuous academic development, student feedback, and the lecturers' portfolios. Each subset of data was classified separately with a specific classifier algorithm. A neural network model was designed to classify the student feedback. A naïve Bayes classifier was used to classify the continuous academic development, and the last data subset, i.e., lecturers' portfolios, was classified via a support vector machine. The results of the data subsets were combined to produce the outcome (an input to another neural network model). Finally, a punished or awarded notification was applied to the lectures. It was concluded that classifying the data as separate datasets did not have a positive indication. The researchers recommended combining the sub-datasets and using one classification algorithm for the system.

The research study in [18] used the same data for the same purpose in a more productive way and improved the accuracy of the recognition system through using a back-propagation neural network with particle swarm optimization. The datasets were first collected and then pre-processed. The most relevant features were identified by using correlation-based feature subset selection and then were fed to the proposed network. The best optimized weights and biases were found by training the neural network via particle swarm optimization. They found that the second proposed study provided a system that had a better accuracy rate than the first.

In [19], a decision tree, neural network, nearest neighbor, and naïve Bayes classifier were used to forecast dropouts in an online program. A 10-fold cross-validation was used. It was concluded that the accuracy rates for the algorithms decision tree, nearest neighbor, neural network, and naive Bayes classifier were 79.7%, 87%, 76.8%, and 73.9%, respectively. In [20], three different classification algorithms—namely, naïve Bayes, C4.5, and ID3—were used to assess the final grades of students who completed the C++ programing language course at the University of Yarmouk in Jordan. The researchers found that the decision tree model outperformed the other models.

**The state of the art of the grey wolf optimizer**

A combination of a support vector machine and the grey wolf optimization (GWO-SVM) approach was presented in [21] to classify the water pollution degree depending on microscopic images of fish liver. GWO-SVM was used for optimizing the parameters. The approach produced better classification accuracy than the standard SVM. The research work concluded that the accuracy increased for each kernel function when training images increased for all classes. The overall performance accuracy of the GWO-SVM was 95.41%.

In [22], a substantial research work was carried out on bioinformatics for the classification of cancer. In this work, a decision tree combined with the Grey Wolf Optimizer approach was presented to choose a small number of valuable genes from an abundance of genes for categorizing cancer. The approach was compared with other classifiers such as Back Propagation Neural Network, Self-Organizing Map, Support Vector Machine, C4.5 and a combined Particle Swarm Optimization with C4.5. They were all applied to cancer datasets of 10 gene expression processes. Their approach outperformed the above-mentioned techniques.

In [23], a system for attribute reduction was proposed based on multi-objective grey wolf optimization. The proposed method tolerates the problems that are common on both wrapper-based feature selection as well as filter-based ones. Grey Wolf Optimization was assessed against Particle Swarm Optimization and a Genetic Algorithm. Their results proved that the GWO produced better results in terms of obtaining global minima.

In [24], a standard neural network trained using the Grey Wolf algorithm was used for categorizing a sonar dataset. The research stated that the GWO had a tremendous ability for resolving higher dimension issues. Their approach was assessed against the Particle Swarm Optimization algorithm, the Gravitational Search Algorithm and the hybrid algorithm of the Particle Swarm Optimization and Gravitational Search. Three types of datasets were used. The comparison was done in terms of the convergence speed, the possibility of trapping in local minima and classification accuracy. Their proposed approach, in most tests, performed better than the other approaches.

In [25], the Grey Wolf Optimization algorithm was used to train the Elman Neural Network for classification and prediction purposes. Two datasets, i.e., Mackey Glass and Breast Cancer, were used in the experiments for gauging their approach. Five different metaheuristic techniques were used in their assessment. Their results showed that the GWO-ENN model generated a better generalization performance.

In [26], a modified version of the Grey Wolf Optimization algorithm was presented to tackle the planning problem of transmission network expansion. This is a significant and difficult problem as it essentially needs to satisfy the load demand in a cost-effective way. The modified GWO was established, gauged and utilized to deal with the transmission network expansion planning issue for Graver's six-bus and Brazilian 46-bus systems. The modified version of the GWO outperformed the other advanced algorithms in terms of accuracy and ability.

## Preliminaries

In this research work, a Grey Wolf Optimizer algorithm was modified. Then, this modified version was applied for optimizing the weight and bias of a modified recurrent neural network to predict student performance. Details about both the standard Grey Wolf Optimizer and Recurrent Neural Networks are first explained.

## Grey wolf optimizer

The Grey Wolf Optimizer (GWO) was first established by Mirjalili in [27]. A swarm-based metaheuristic algorithm is inspired by the behaviors of the Grey Wolf. Thus, it is a nature-inspired algorithm that mimics a mechanism in nature, such as particle swarm optimization (PSO) [28] (derived from bird and fish behaviors), ant colony optimization (ACO) [29] (which depends on the behavior of ant colonies), and the bees algorithm (BA) (drawn from the food foraging behavior of honey bees) [30]. These algorithms are considered to be very useful due to their speed, simplicity, and faster convergence in finding a global optimum solution in comparison with deterministic methods.

The algorithm is motivated by the grey wolves' hunting style. This algorithm divided grey wolves into four different groups: Alpha (α), Beta (β), Delta (δ), and Omega (ω). The first three (Alpha, Beta, and Delta) are known as the three finest fitting wolves. These three wolves will direct omega wolves to favorable zones in the search area. The positions of wolves are changed during optimization around alpha, beta, and delta via the following equations (1) and (2):

$$\vec{D} = |\vec{X}p(t) \cdot \vec{C} - \vec{X}(t)| \qquad (1)$$

$$\vec{X}(t+1) = \vec{X}p(t) - \vec{D} \cdot \vec{A} \qquad (2)$$

where vector $\vec{D}$ represents the difference between the position of the prey and predator that is computed, t denotes the current iteration, vector $\vec{X}p$ specifies the prey's position, and vector $\vec{X}$ signifies the grey wolf's position. The vector values of both $\vec{A}$ and $\vec{C}$ can be determined via the following equations:

$$\vec{A} = \vec{r}1a \cdot 2a \qquad (3)$$
$$\vec{C} = \vec{r}2 \cdot 2 \qquad (4)$$

where a can be decreased linearly starting from 2 down to zero and both vectors $\vec{r}1$ and $\vec{r}2$ are random values between 0 and 1.

Notice that the notion of updating positions through equations (1) and (2) is demonstrated in Fig. 1. Note that a wolf might change its position relative to its prey in the position of (X, Y).

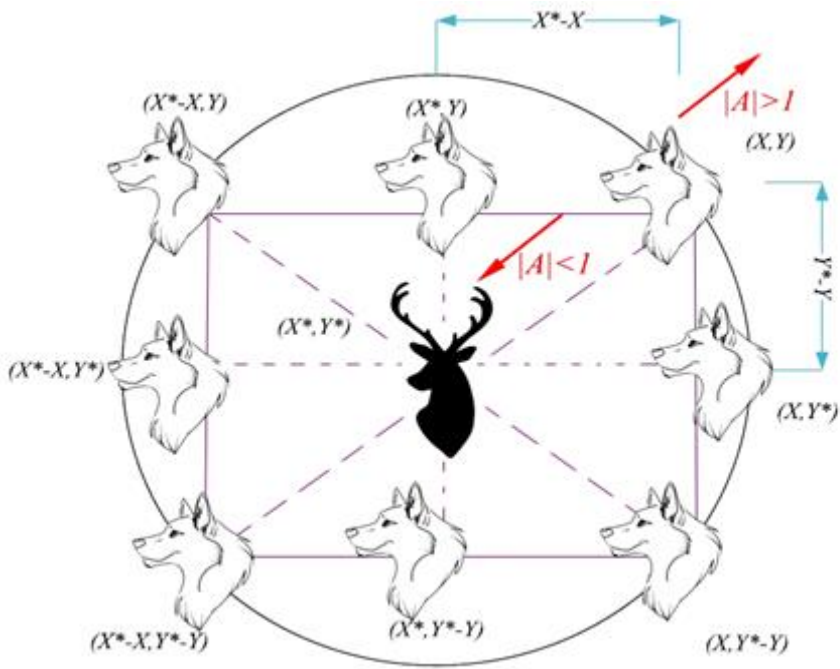

**Fig. 1. Search agents position updating mechanism and effects of A on it.**

The position of the prey or the best three solutions in a GWO algorithm are constantly expected to be alpha, beta, and delta, in that order, during optimization. The other wolves are called omegas; they can change their positions towards alpha, beta, and delta.

The positions of the omega wolves are updated via the following equations. The equations compute the approximate distance between the alpha, beta, and delta wolves and the current solution, respectively [27]:

$$\vec{D}\alpha = |\vec{X}\alpha \cdot \vec{C}1 - \vec{X}| \quad (5)$$

$$\vec{D}\beta = |\vec{X}\beta \cdot \vec{C}2 - \vec{X}| \quad (6)$$

$$\vec{D}\delta = |\vec{X}\delta \cdot \vec{C}3 - \vec{X}| \quad (7)$$

where the values of vectors $\vec{C}1$, $\vec{C}2$, and $\vec{C}3$ are set randomly; $\vec{X}\alpha$, $\vec{X}\beta$, and $\vec{X}\delta$ are the positions of alpha, beta, and delta, respectively; and $\vec{X}$ is the position of the current solution. The step sizes of the omega wolves towards alpha, beta and delta are defined via (5), (6) and (7).

The final position of the current solution is calculated when the distances have been described as follows:

$$\vec{X}1 = \vec{X}\alpha - (\vec{D}\alpha) \cdot \vec{A}1 \quad (8)$$

$$\vec{X}2 = \vec{X}\beta - (\vec{D}\beta) \cdot \vec{A}2 \quad (9)$$

$$\vec{X}3 = \vec{X}\delta - (\vec{D}\delta) \cdot \vec{A}3 \quad (10)$$

$$\vec{X}(t+1) = \frac{(\vec{X}1 + \vec{X}2 + \vec{X}3)}{3} \quad (11)$$

where $\vec{A}1, \vec{A}2$, and $\vec{A}3$ represent random vectors.

The random and adaptive vectors $\vec{A}$ and $\vec{C}$ provide both exploration and exploitation for the algorithm, as shown in (Fig. 1). As can be seen, the exploration occurs if $|A| > 1$ or $|A| < -1$. The exploration is also facilitated by vector $\vec{C}$ if it is greater than 1. However, if $\vec{A}$ is smaller than 1 and $\vec{C}$ is smaller than 1, then the exploitation occurs.

A suitable technique is suggested in the algorithm to solve the entrapment of local optima. Thus, to emphasize exploitation, it is noticed during optimization that as the iteration counter increases, A decreases linearly. However, C is randomly produced during the optimization to emphasize exploration/exploitation at any stage.

The GWO Algorithm's pseudo code can be expressed as follows:

*Initialize the grey wolf population Xi, where* $i = 1, 2, 3, 4 \ldots n$
*Initialize* $a, A,$ *and* $C$
*the fitness of each search agent is computed*
$X\alpha$, *is the first finest search agent*
$X\beta$, *is the second finest search agent*
$X\delta$, *is the third finest search agent*
**While** (*iteration* < *Maximum iteration number*)
**for** *each search agent*
*Modify the current search agent's position via equation* (11)
**end for**
*Modify* $A, C, \& a$
*the fitness for all search agents is computed*
*Modify* $X\alpha, X\beta, \& X\delta$
*iteration* = *iteration* + 1
**end While**
*return* $X\alpha$

## Recurrent neural network

The Multi-Layer Perceptron feeds data from lower layers to higher layers, whereas recurrent neural networks (RNNs) are considered bi-directional data flow neural networks. The data flow propagates from previous processing phases to earlier phases. In this research work, the concept of a simple recurrent neural is used, which was first proposed by Jeff Elman [31].

The model in Fig. 2 uses a three-layered network. At the hidden layer, the output from each hidden neuron at time $(t - 1)$ is saved in context neurons and then, at time $(t)$, is fed together with the initial input to the hidden layer. Thus, copies of the previous values of the hidden neurons are continuously kept by the context neurons, due to the propagation through the recurrent connections from time$(t-1)$, before a parameter-updating rule is applied at time$(t)$. Consequently, the network model keeps and acquires a set of state summarizing previous inputs.

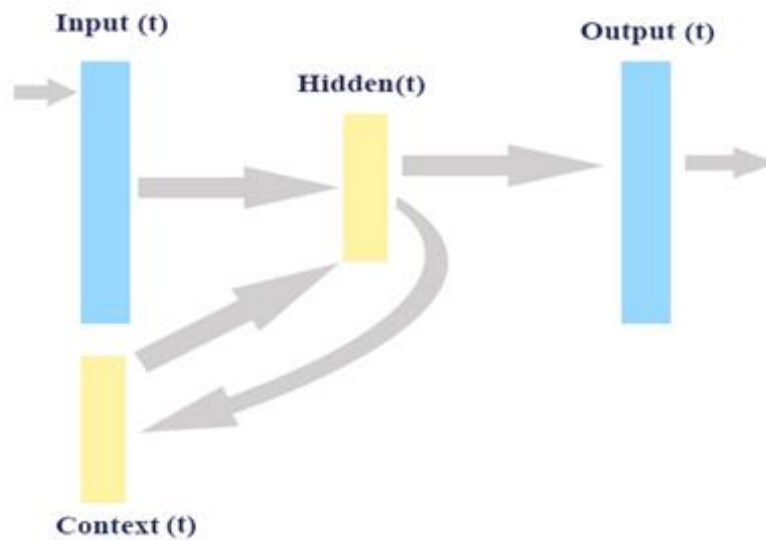

**Fig. 2. A simple RNN model.**

## Student dataset description

In most universities in Kurdistan, students are registered in a general English course in their first academic year. The system presented in this study forecasts the students' outcome in the course and categorizes them as either passing or failing students through a modified recurrent neural network. The raw data were collected from Salahaddin University-Erbil, College of Engineering [32]. The data consist of questionnaires and student documents and are used in this research to classify the students. The information in the datasets includes the students' past achievements, social settings and academic environments. This study principally focuses on the socio-economic background and the tutors' expertise. The features and descriptions of the dataset and the implementation codes of all models can be found via the following link: "https://github.com/Tarik4Rashid4/student-performance".

# Research methodology

A modified recurrent neural network with a modified GWO was used for predicting student performance. This research improves on the previous study on student academic performance in [32]. The problems of back-propagation have been highlighted and the data have been collected from our previous research work about student performance in English courses at the College of Engineering at Salahaddin University. The data consist of 287 samples. In this proposed approach, an RNN model is developed by using the modified GWO to optimize the values of biases and weights of the model. Initially, the neural network model is trained by using a training dataset, and its weights and biases are optimized by using a modified recurrent network with GWO. In the second step, to evaluate the trained model, the designed model is tested with a predefined testing dataset. For the validation procedure, cross validation of 5-fold is used for attaining high accuracy and performance. In this study, MATLAB is used for the implementation. The key stages of this work are explained below:

## Preprocessing

After data collection using questionnaires, the dataset is normalized for preparation and processing. Since cross validation is used, the data arrangement in a structure consists of five sets to 5-folds. Each set contains an equal number of passing and failing students.

## Feature selection

This is an important stage for classification. The most relevant features for classification are selected. In other words, features that have no contribution to the classification output are removed. The Correlation Attribute Evaluation in Weka is used to evaluate the features by calculating the correlation between the class and features. Through use of the algorithm, it was determined that features such as the College and the Address of the High School for the Town and Village had no effect on the output. Consequently, College and High School (Village) are eliminated from the features. To conclude, our dataset consisted of 18 input features and one output feature.

## Classification

There are several conventional classification algorithms in the educational data mining field. A modified recurrent neural network with a modified GWO is redeveloped for the classification of student outcomes in a course. It is a two-step procedure. The modifications are conducted on both the RNN and GWO to form a new model called the Modified Recurrent Neural Networks with Grey Wolf Optimizer (M-RNNGWO) model. Details of the modifications are given in the next three subsections:

**A modified grey wolf optimizer**

In this research work, a variant of the GWO is produced by adding two simple modifications to the original GWO algorithm to optimize the parameters of the modified recurrent neural network to classify students. The outcomes demonstrate that the modifications positively affected the classification accuracy. As mentioned in the above, the GWP algorithm divided the population into four sets, i.e., Alpha (α), Beta (β), Delta (δ), and Omega (ω). Alpha, Beta, and Delta are recognized as the three fittest wolves (or best solutions) that direct the Omega wolves on how to achieve the optimal search space area. The first modification to this model is to add another best solution to Alpha, Beta, and Delta, called Gamma (see equation (12). When the Omega wolves update their positions with respect to the best positions, in this case, they (Omega wolves) update their positions with more best positions (Alpha, Beta, Delta and Gamma) than the standard algorithm (GWO). The second modification involves defining the step size of the omega wolves (which moves from Alpha, Beta, Delta and Gamma, correspondingly), as shown in equations (8), (9), (10), and (11). The variables $\vec{X}1, \vec{X}2, \vec{X}3,$ and $\vec{X}4$ are calculated instead of using the Alpha, Beta, Delta, and Gamma distances ($\vec{D}\alpha, \vec{D}\beta, \vec{D}\delta,$ and $\vec{D}\gamma$ are found by equations (5), (6), and (7) individually). The average of these distances is taken as shown in equation (13):

$$\vec{D}\gamma = |\vec{X}\gamma \cdot \vec{C}4 - \vec{X}| \qquad (12)$$

where $\vec{D}\gamma$ is the approximate distance between Gamma and the current solution and $\vec{C}4$ is a random vector. The value of $\vec{C}$ was defined above in the GWO Algorithm. $\vec{X}\gamma$ Shows the position of Gamma, and $\vec{X}$ is the position of the current solution,

$$\vec{D}avg = \frac{\vec{D}\alpha + \vec{D}\beta + \vec{D}\delta + \vec{D}\gamma}{\# \ of \ 4 \ best \ solutions} \qquad (13)$$

where $\vec{D}avg$ denotes the average of the approximate distances between Alpha, Beta, Delta, and Gamma and the current solution, individually. Then, equations (6), (7), and (8) will be updated as follows:

$$\vec{X}1 = \vec{X}\alpha - (\vec{D}avg) \cdot \vec{A}1 \quad (14)$$

$$\vec{X}2 = \vec{X}\beta - (\vec{D}avg) \cdot \vec{A}2 \quad (15)$$

$$\vec{X}3 = \vec{X}\delta - (\vec{D}avg) \cdot \vec{A}3 \quad (16)$$

where $\vec{A}1, \vec{A}2$, and $\vec{A}3$ denote random vectors. The value of $\vec{A}$ is defined above in the GWO Algorithm.

Furthermore, another equation will be expressed before calculating the current solution's final position as follows:

$$\vec{X}4 = \vec{X}\gamma - (\vec{D}avg) \cdot \vec{A}4 \quad (17)$$

where $\vec{A}4$ denotes a random vector. Finally, to calculate the current solution's final position, we update equation (11) as follows:

$$\vec{X}(t+1) = \frac{(\vec{X}1 + \vec{X}2 + \vec{X}3 + \vec{X}4)}{\#\ of\ 4\ best\ solutions} \quad (18)$$

**A modified RNN**

The developed neural network model consists of using the concept of RNN on a multilayer perceptron with two hidden layers and two context layers (one context for each hidden layer). The structure of the model is as follows: 18,10-10,10-10,1; 18 neurons in the input layer, 10 neurons in the first hidden layer with 10 neurons for the first context layer, 10 neurons in the second hidden layer with 10 neurons for the second context layer, and 1 neuron in the output layer. The neurons of the first and second context layers are copies of neurons from the previous time of the first and second hidden layers, respectively (see equations below).

$$C_i^1(t) = h_j^1(t-1) \quad (19)$$

$C_l^1(t)$ represents the lth neuron in the first context layer at time t, or it is equal to $h_j^1(t-1)$, which represents the jth neuron in the first hidden layer at the previous time.

$$C_m^2(t) = h_g^2(t-1) \tag{20}$$

$C_m^2(t)$ represents the lth neuron in the second context layer at time t or it equals to $h_g^2(t-1)$, which represents the jth neuron in the second hidden layer at the previous time.

The feed-forward to the first hidden layer can be stated as follows:

$$h_j^1(t) = f\left(\sum_i^I v_{ij}^1 x_i(t)\right) + f\left(\sum_l^{Con^1} u_{lj}^1 C_l^1(t)\right) \tag{21}$$

$$f(net) = \frac{1}{1+e^{-net}} \tag{22}$$

where f(net) represents an activation function in which both Sigmoid and Softamax are used for experimental purposes in each hidden neuron at the hidden layers. $v_{ij}^1$, and $u_{lj}^1$, indicate weight connections concerning the first hidden layer $h_j^1(t)$ and the input layer $x_i(t)$, and the first hidden layer $h_j^1(t)$ and the first context layer $C_l^1(t)$, respectively.

The feed-forward to the second hidden layer can be stated as follows:

$$h_g^2(t) = f\left(\sum_j^{H1} v_{jg}^2 h_j^1(t)\right) + f\left(\sum_m^{Con^2} u_{mg}^2 C_m^2(t)\right) \tag{23}$$

where $v_{jg}^2$ and $u_{mg}^2$ indicate weight connections between the second hidden layer $h_g^2(t)$ and the first layer $h_j^1(t)$, and between the second hidden layer $h_g^2(t)$ and the second context layer $C_m^2(t)$, respectively.

The feed-forward to the output layer can be written as follows:

$$O_k(t) = f\sum_g^{H2} w_{gk} h_g^2(t) \tag{24}$$

where $w_{gk}$ represents the weight connection between the output layer $O_k(t)$ and the second hidden layer $h_g^2(t)$.

Also, the objective function here for training the model is the least Mean Square Error (MSE) to obtain the highest classification, where MSE represents the variance between the predicted output in the form of the improved RNN with GWO and the target output. The MSE is calculated as follows:

$$\text{MSE}_p = \frac{1}{n}\sum_{k=1}^{n}(O_k(t) - d_k(t))^2 \qquad (25)$$

where n represents the number of output neurons and $d_k(t)$ and $O_j^k$ denote the desired and the actual outputs of the $k^{th}$ neuron. The total MSE across all samples can be expressed as follows:

$$\text{Total}_{MSE} = \frac{1}{n}\sum_{p=1}^{S}\text{MSE}_p \qquad (26)$$

where p represents a sample pattern and S represents the number of training patterns. Notice that the input to the modified GWO is the MSE and that the output is weights and biases.

## The M-RNNGWO

In this paper, a 5-fold cross validation method is used for verification of the classification. In each fold, the training step is processed as shown in Fig. 3.

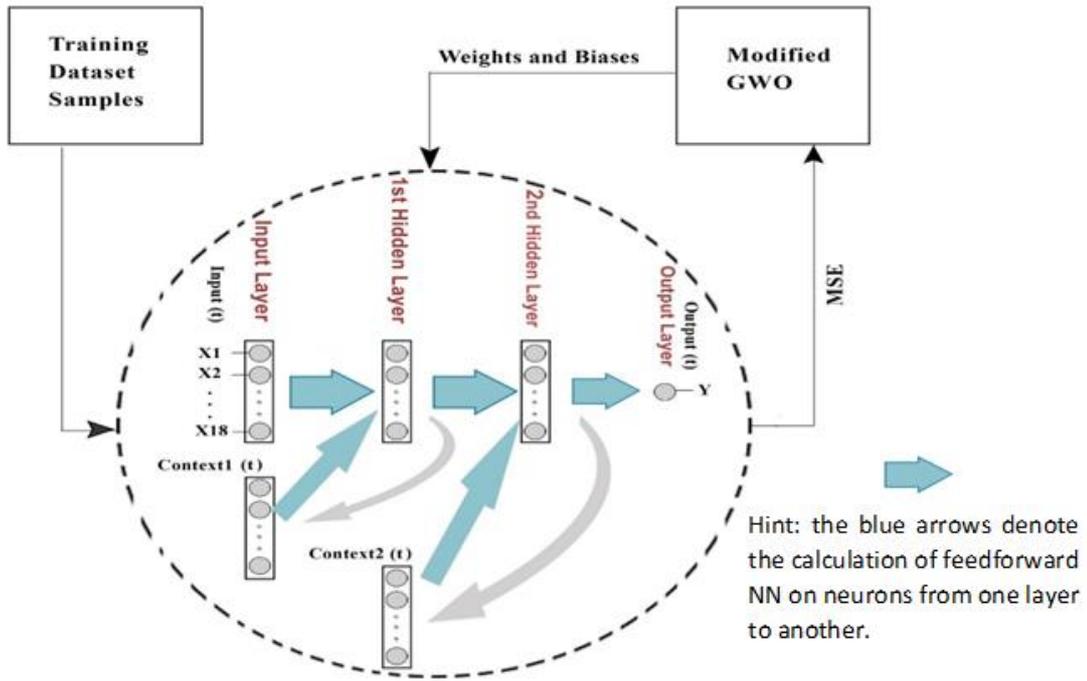

**Fig. 3. Training the modified RNN through the modified GWO (M-RNNGWO).**

In the training phase, the M-RNNGWO has two parts: the modified RNN and the modified GWO. The modified GWO initially sets its variables and weights and biases to the modified RNN in the form of a vector. Thus, the vector contains values that represent a weight or a bias in the M-RNNGWO. The first sample is then fed to the modified RNN, which is followed by a copy of the output from the first hidden layer at time (t) being held in the first context layer. Next, at time t+1, the net is fed back to the first hidden layer. Simultaneously, a copy of the output from the second hidden layer at time (t) is held in the second context layer. Then, at time t+1, the net is fed back to the second hidden layer. This model of the recurrent neural network preserves and learns a set of state summarizing previous inputs. This process continues iteratively to feed all the other training samples to the modified RNN using the same initialized weights and biases. After computing the $Total_{MSE}$ over the training samples, then the modified GWO receives the $Total_{MSE}$. The modified GWO assesses the $Total_{MSE}$ with fitness around the four best wolves, i.e., alpha, beta, delta, and gamma. Then, after the fitness and the position of each of the best wolves are modernized, the vector of weights and biases, which denotes the positions of the search agents, is adjusted iteratively based on the number of search agents with respect to alpha, beta, delta, and gamma. After the weights and biases are updated by the modified GWO, then they are passed to the modified RNN. In conclusion, the training samples and the updated weights and biases are used to train the modified RNN to archive a new $\text{Total}_{\text{MSE}}$. The training procedure is constant until the termination condition is met. To finish, the optimized weights and biases are used to test the M-RNNGWO by using a testing dataset without using a modified GWO.

**Weight Complexity Computation**

In all models, the user is able to specify hidden layers, context layers, and neurons at each layer. The basic exercise is to choose the fewest of the above parameters possible to find the best feasible arrangement per the requirements. However, practically, this does not come easily as there have to be more trials via using various structures and gauging their results to determine the best fitting model structure to deal with the task. Based on our trials, one or two hidden layers can be sufficient. The following equation defines the connection weights computation for M-RNNGWO:

$$d = (i + 1) * h1 + (h1 + 1) * h2 + (h2 + 1) * o + c1 * h1 + c2 * h2 \qquad (27)$$

where $d$ denotes the dimension of the problem, $i, h1, h2, c1, c2, and\ o$ represent the neurons at the input layer, the neurons at the first hidden layer, the neurons at the second hidden layer, the neurons at the first context, the neurons at the second context, and the neurons at the output, respectively. Both the input and the hidden layers have a bias; thus, a neuron is added to each.

# Results and discussion

The results of the classification using cross validation are shown in Table 1. The dataset was divided into five groups (5-folds), named as X1, X2, X3, X4, and X5. The first three groups consisted of 57 samples, and the last two group contained 58 samples. In each fold run, four groups were fed to the network model, as the training dataset consisted of approximately 230 samples, and the remaining were rolled, as the testing dataset consisted of approximately 57 samples to test the network. The results showed that the training classification rates in the folds were 99.56%, 99.56%, 99.56%, 99.12%, and 99.56%, and the average rate was 99.47%. Also, the classification rates for the testing phase for each fold were 96.49%, 100%, 100%, 98.27%, and 98.27%, and the average was 98.60%. It can be seen from the results that when a smaller $\text{Total}_{\text{MSE}}$ is produced, a better classification rate is obtained. For example, in Fold 1, the classification rate in the testing phase is 96.49% and its Total$_{MSE}$ is 0.009, but when the testing rate is 100.00% in the second and third folds, then the total MSE is 0.002, which is a smaller MSE.

**Table 1. Shows classification results.**

| Fold No. | Training/ Testing | Dataset | No. of Samples | MSE | Classification Rate |
|---|---|---|---|---|---|
| Fold 1 | Training | X2+X3+X4+X5 | 230 | 0.0029211 | 99.56% |
|  | Testing | X1 | 57 | 0.0091276 | 96.49% |
| Fold 2 | Training | X1+X3+X4+X5 | 230 | 0.002923 | 99.56% |
|  | Testing | X2 | 57 | 0.0027236 | 100.00% |
| Fold 3 | Training | X1+X2+X4+X5 | 230 | 0.0030488 | 99.56% |
|  | Testing | X3 | 57 | 0.0027124 | 100.00% |
| Fold 4 | Training | X1+X2+X3+X5 | 229 | 0.0030902 | 99.12% |
|  | Testing | X4 | 58 | 0.0030253 | 98.27% |
| Fold 5 | Training | X1+X2+X3+X4 | 229 | 0.0028664 | 99.56% |
|  | Testing | X5 | 58 | 0.0033664 | 98.27% |
| Average | Training |  |  | 0.0029699 | 99.47% |
|  | Testing |  |  | 0.00419106 | 98.60% |

The performance and outcomes of the students are shown in Table 2. It shows that there was a total of 287 students. The total number of students who passed the course was 183, and the total number of students who failed was 104. In the first run, from the passing students, 36 students were classified successfully out of 37 students, with a success rate of 97.29%, and 1 student was not correctly classified. Of the failed students, 19 students were classified successfully out of 20, with the success rate of 95.00%, and 1 student was not correctly classified. In the second and third folds, all 37 students that passed were classified successfully. All 20 students who failed were also properly classified. In Folds 4 and 5, 35 of 36 students who passed were classified correctly, resulting in a success rate of 97.22%. All 22 failing students were classified correctly as well.

In addition to the folds, 180 students were correctly classified out of 183 passing students, with a success rate of 98.34%. Of the students who failed, 103 were classified successfully out of 104, with a success rate of 99.00%.

**Table 2. Performance and outcomes of the students.**

| Fold No. | Dataset | No. of Samples | Passing Students | | | Failing Students | | |
|---|---|---|---|---|---|---|---|---|
| | | | No. of Students | No. of Correctly Classified Students | Success Rate | No. of Students | No. of Correctly Classified Students | Success Rate |
| Fold 1 | X1 | 57 | 37 | 36 | 97.29% | 20 | 19 | 95.00% |
| Fold 2 | X2 | 57 | 37 | 37 | 100.00% | 20 | 20 | 100.00% |
| Fold 3 | X3 | 57 | 37 | 37 | 100.00% | 20 | 20 | 100.00% |
| Fold 4 | X4 | 58 | 36 | 35 | 97.22% | 22 | 22 | 100.00% |
| Fold 5 | X5 | 58 | 36 | 35 | 97.22% | 22 | 22 | 100.00% |
| Total | | 287 | 183 | 180 | | 104 | 103 | |
| Average | | | | | 98.34% | | | 99.00% |

Fig. 4 shows that the M-RNNGWO obtained the best accuracy among the other methods. The M-RNNGWO was evaluated against some other techniques. The M-RNNGWO was compared to a standard GWO with RNN (RNNGWO). The M-RNNGWO produced 98% accuracy, while the RNNGWO produced 94% accuracy. The modified GWO with Multilayer Perceptron (M-MLPGWO) obtained 88% accuracy, and the standard GWO with Multilayer Perceptron (MLPGWO) obtained 77% accuracy. The GWO with Cascade MLP (CMLPGWO) produced 89% accuracy. However, the modified GWO with Cascade MLP (M-CMLPGWO) produced 84 % accuracty. The accuracies of the other algorithms are as follows: Backpropagation Neural Network (BPNN), 76%; Naïve Bayes Classifier, 73%; and Random Forest, 81%.

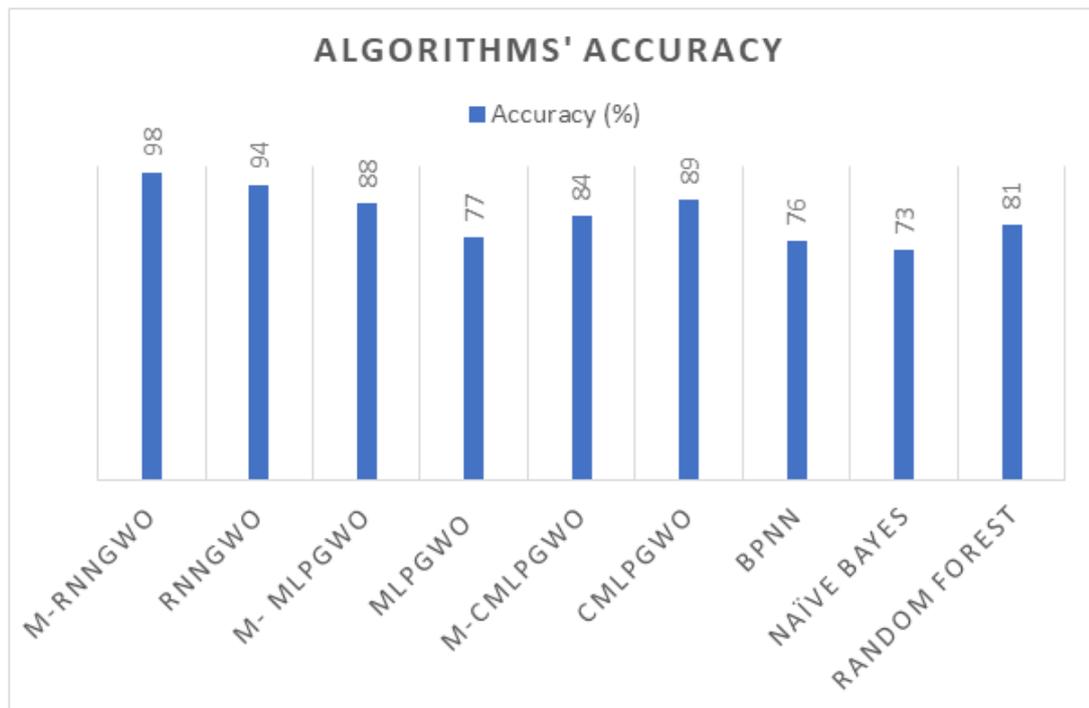

**Fig. 4. Accuracy of the algorithms.**

A confusion matrix is used as another measurement in the proposed classification techniques to gauge the students' classification results. The testing results for the M-RNNGWO are assessed in the following discussion.

Table 3 demonstrates the confusion matrix in the first fold for the M-RNNGWO. The predicted number of true positives (passed) and the predicted number of false negatives (failed) were 36 and 1, respectively, the predicted number of false positives (passed) and the predicted number of true negatives (failed) were 1 and 19, respectively.

**Table 3. Confusion matrix for M-RNNGWO – fold (1).**

|  |  | Predicted | |
|---|---|---|---|
|  |  | **Passed** | **Failed** |
| **Actual class** | **Passed** | 36 | 1 |
|  | **Failed** | 1 | 19 |

Notice from the above table that the Sensitivity or True Positive Rate (TPR), Specificity or True Negative Rate (TNR) and the Positive Predictive Value (PPV) or Precision can be computed. The Positive Predictive Value (PPV) or Precision governs the success rate in passing students, whereas the Negative Predictive Value (NPV) governs the success rate in failing students. Likewise, the accuracy of the network can also be computed. Detailed descriptions of computing the above variables are explained via equations (28), (29), (30), (31), and (32), as follows:

$$\text{Sensitivity} = \frac{36}{36+1} = 0.97 \quad (28)$$

$$\text{Specificity} = \frac{19}{19+1} = 0.95 \quad (29)$$

$$\text{PPV} = \frac{36}{36+1} = 0.97 \quad (30)$$

$$\text{NPV} = \frac{19}{19+1} = 0.95 \quad (31)$$

$$\text{Accuracy} = \frac{36+19}{36+19+1+1} = 0.96 \quad (32)$$

Notice from the above computations that the sensitivity value was 0.97 indicating that the TPR was 97%, the specificity value was 0.95 indicating that the TNR was 95%, the PPV was 0.97 indicating that the success rate in passing students was 97%, the NPV was 0.95 indicating that the success rate in failing students was 95%, and the obtained accuracy of the network in the first fold was 96%.

Table 4 demonstrates the results of the other folds. The table contains the computation of the confusion matrix for M-RNNGWO generally.

**Table 4. Evaluation of the confusion matrix.**

| Fold No. | Sensitivity | Specificity | PPV | NPV | Accuracy |
|---|---|---|---|---|---|
| Fold 1 | 0.97 | 0.95 | 0.97 | 0.95 | 0.9649 |
| Fold 2 | 1.00 | 1.00 | 1.00 | 1.00 | 1.00 |
| Fold 3 | 1.00 | 1.00 | 1.00 | 1.00 | 1.00 |
| Fold 4 | 1.00 | 0.95 | 0.97 | 1.00 | 0.9827 |
| Fold 5 | 1.00 | 0.95 | 0.97 | 1.00 | 0.9827 |
| Average | 0.99 | 0.97 | 0.98 | 0.99 | 0.9860 |

Table 5, highlights the dimension of the problem for the proposed model compared with the other models. We can see that the RNN outperforms the other neural network types. Whenever we use the M-RNNGWO, the accuracy is greater than the one that uses the RNNGWO. In the algorithms, we used two hidden layers for the RNN and one hidden layer for the other neural network algorithms. There is another feature that makes the RNN outperform other neural network models, which is the dimension of the problem or the number of connections. These connections are assigned as the positions of the wolves in the GWO. The GWO with the least number of positions updates its positions faster than the one with a greater number of positions since it needs less time to update the positions. Therefore, the RNN finishes the process earlier than the other used neural network types.

**Table 5. Weight complexity computation of the models.**

| Algorithm | No. Connections | Search Agents No. | No. Iteration(s) | No. Hidden Layers | Testing Rate |
|---|---|---|---|---|---|
| **M-RNNGWO** | 511 | 50 | 75 | 2 | 98.60% |
| **RNNGWO** | 511 | 50 | 75 | 2 | 94.40% |
| **M-MLP GWO** | 528 | 50 | 75 | 1 | 88.00% |
| **MLPGWO** | 528 | 50 | 75 | 1 | 77.05% |
| **M-CMLPGWO** | 544 | 50 | 75 | 1 | 84.40% |
| **CMLPGWO** | 544 | 50 | 75 | 1 | 89.35% |

In addition to the above results, further statistical experiments on the proposed models are conduced to evaluate the obtained results. Thus, the accuracy can be measured by the area under the curve or the area under the ROC curve. Literally, ROC stands for the Receiver Operating Characteristic. The ROC analysis is related to the Signal Detection Theory established in the course of the second world war for analyzing radar images. The radar operators needed to determine whether glitches on the monitor characterized enemy goals, amicable ships, or noise. The theory of signal detection is able to measure the radar receiver operators' ability to detect these substantial differences. This capability is called the Receiver Operating Characteristics. The experiment accuracy relies on how fine the experiment splits the students being tested into those who passed and failed. Fig. 5 (a, b, c, d, e, f) shows the ROC curve for the proposed models. Accuracy is gauged by the area under the ROC curve. For example, an area of 0.5 indicates an insignificant test and 1 indicates a perfect test. Determining the area is very difficult to describe, and it is outside the scope of this paper. Commonly, there are two approaches utilized for determining the area under the curve, i.e., parametric and non-parametric. The parametric approach uses a maximum probability estimator for fitting a flat curve to the data samples, and the non-parametric approach depends on building trapezoids below the curve as an approximation of the area [33, 34]. The M-RNNGWO produced the best area under the ROC curve, with AUC=0.872, and the MLPGWO produced the lowest area under the ROC curve, with AUC=0.657, among the others, as shown in Fig. 5 (a) and (d).

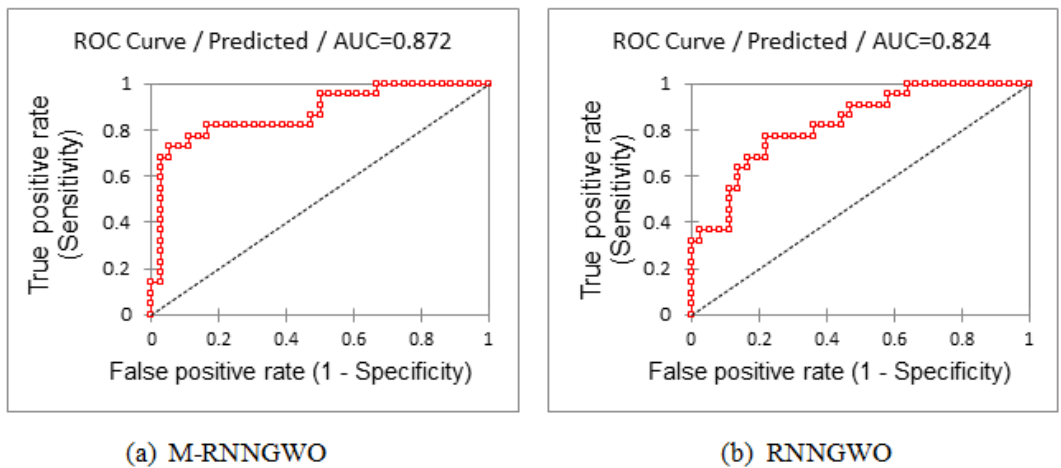

**Fig. 5 (a, b). Shows Area under the ROC curve.**

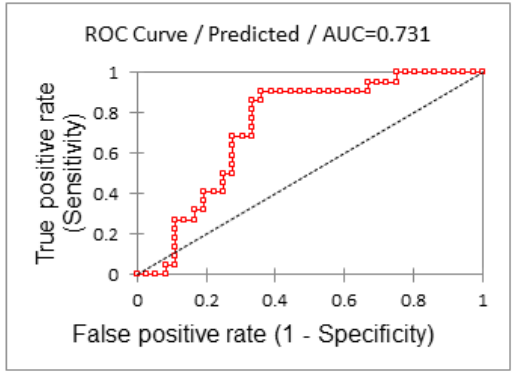
(c) M-MLPGWO

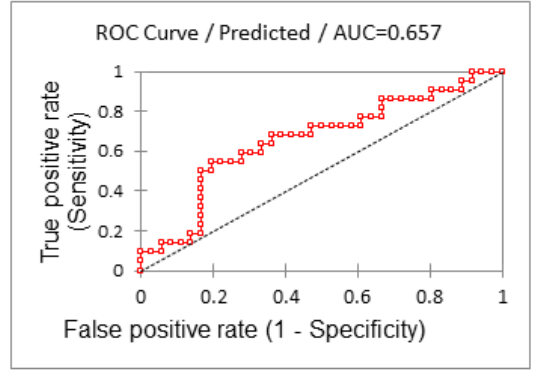
(d) MLPGWO

**Fig. 5 (c, d). Shows Area under the ROC curve.**

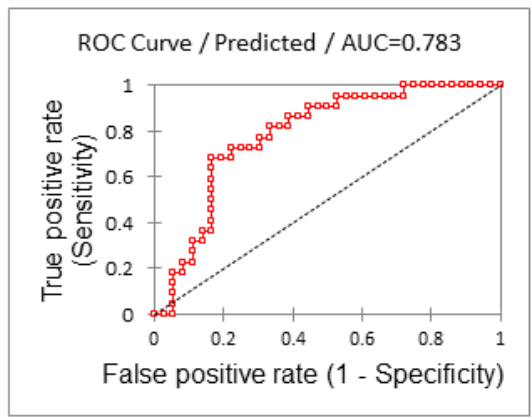
(e) M-CMLPGWO

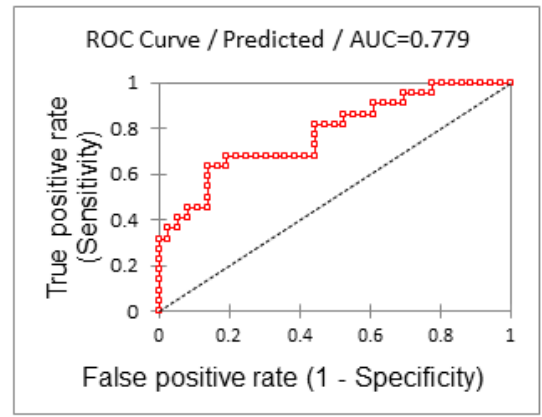
(f) CMLPGWO

**Fig. 5 (e, f). Shows Area under the ROC curve.**

Fig. 6 (a, b, c, d, e, f) is another way of showing these differences among the produced models. The figure shows the sensitivity and specificity against the predicated values for each model.

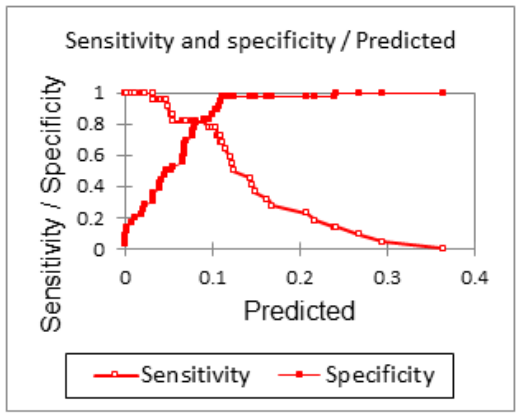 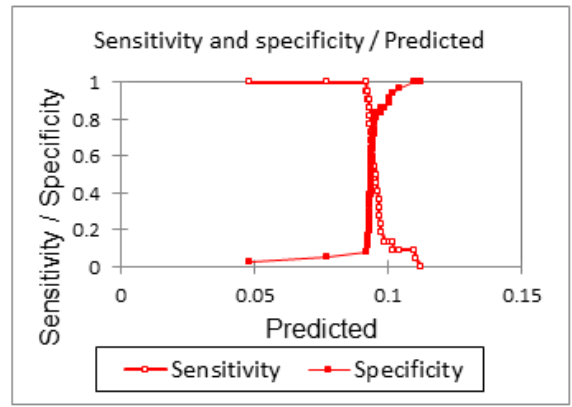

(a) M-RNNGWO        (b) RNNGWO

**Fig. 6 (a, b). Shows the sensitivity and specificity against the predicated values.**

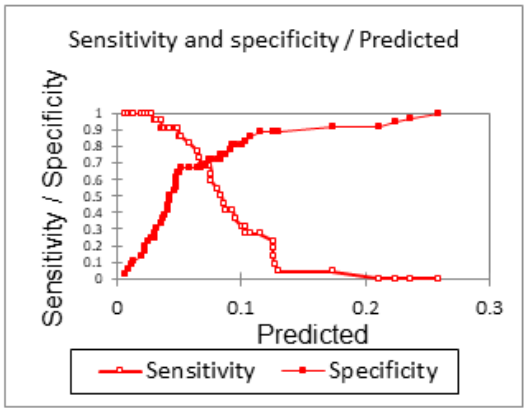 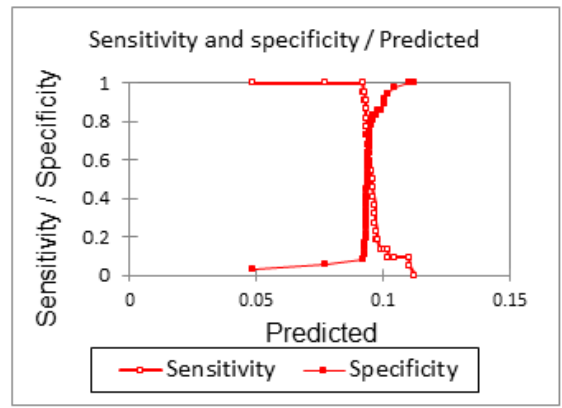

(c) M-MLPGWO        (d) MLPGWO

**Fig. 6 (c, d). Shows the sensitivity and specificity against the predicated values.**

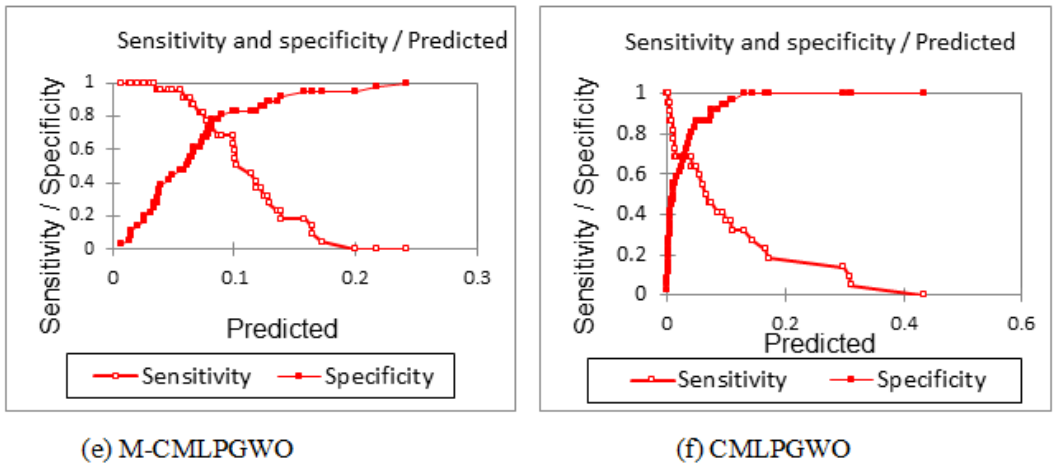

**Fig. 6 (e, f). Shows the sensitivity and specificity against the predicated values.**

Fig. 6 shows a normal ROC graph for a prediction system. As illustrated in the figures the best performance can reach at a high sensitivity of 0.6 to 0.7 for all cases at which the trained classier of the neural network can have a specificity value of less than 0.1, so that the classier might be better used in cases when sensitivity is far more important than the specificity.

In addition, Fig. 7 (a, b, c, d, e, f) shows the True positive, True negative, False positive, and False negative against the Predicted values for all of the proposed models.

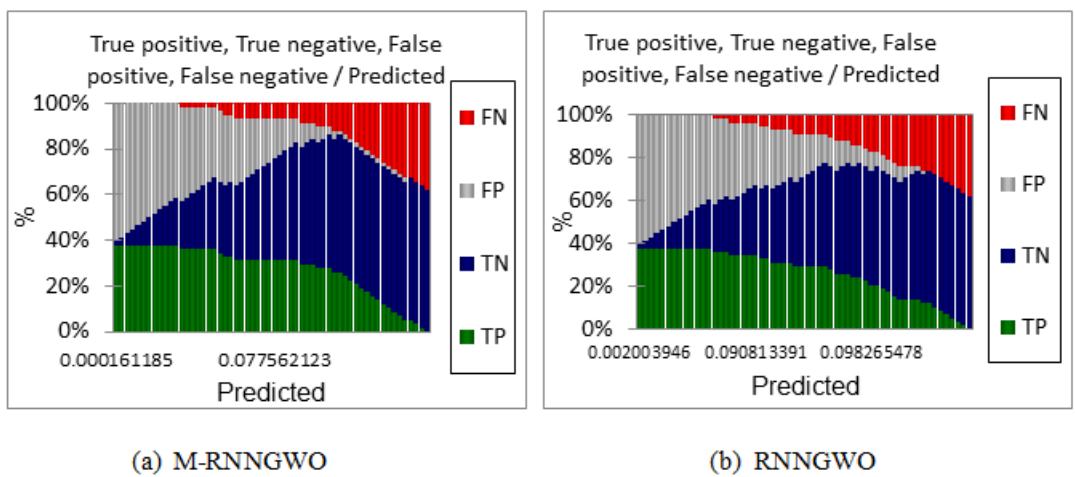

**Fig. 7 (a, b). Show the True positive, True negative, False positive, and False negative against the Predicted values.**

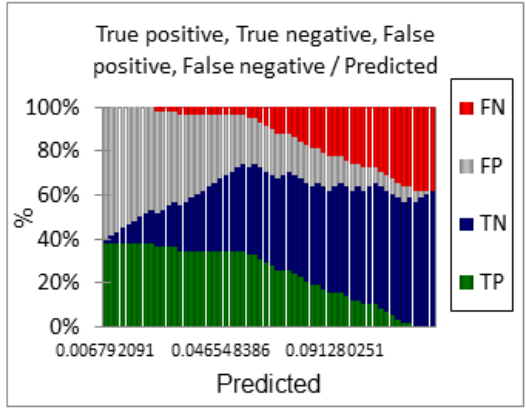
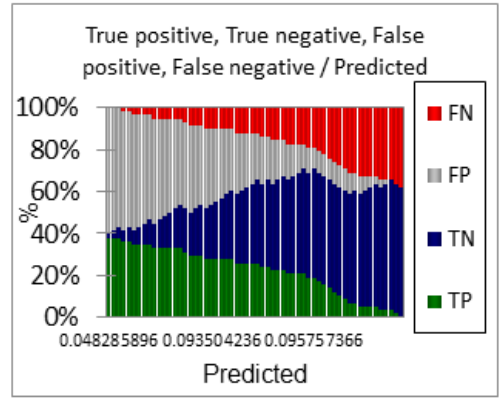

(c) M-MLPGWO    (d) MLPGWO

**Fig. 7 (c, d). Show the True positive, True negative, False positive, and False negative against the Predicted values.**

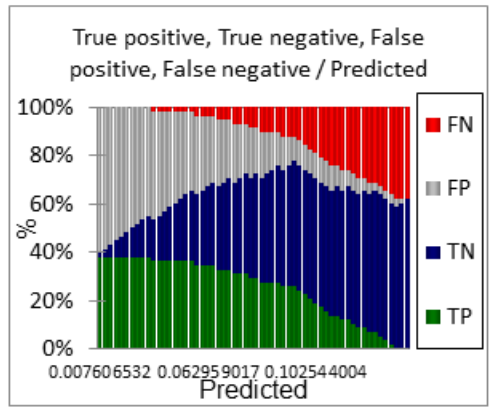
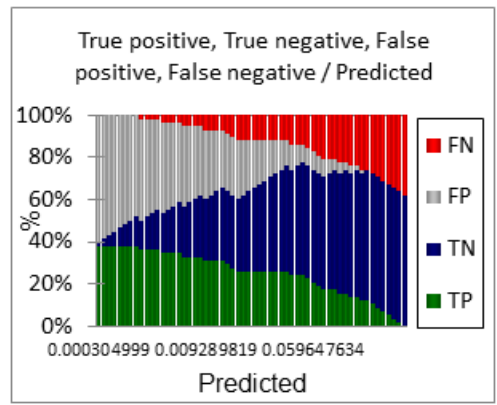

(e) M-CMLPGWO    (f) CMLPGWO

**Fig. 7 (e, f). Show the True positive, True negative, False positive, and False negative against the Predicted values.**

Fig 8 shows the classification efficiency on the underlying data set as the classifier tends to predict positive for the smaller values and mostly negative for the next half. As it can be noticed from the figures the false positive is more prominent and have been predicted by the classier than false negative. This result might be important for most of the practical applications, thus, it needs closer attention for the positive classes.

Finally, the proposed models are also evaluated against two other models, i.e., the Logistic Regression and Elastic Net. For this test, a Weka tool was used to obtain the classification results. Tables 6 and 7 provide details about the performance of both models on the same dataset.

**Table 6. Shows results produced by the logistic regression model.**

| Correctly Classified Instances | Incorrectly Classified Instances | Total number of instances | Classification Rate | MAE | Precision | Recall | ROC Area |
|---|---|---|---|---|---|---|---|
| 43 | 14 | 57 | 75.4386 % | 0.2853 | 0.754 | 0.754 | 0.801 |

**Table 7. Shows results produced by the elastic net model.**

| Correlation coefficient | MAE | RMSE | RAE | RRAE | Total number of instances |
|---|---|---|---|---|---|
| 0.7039 | 0.3358 | 0.3771 | 70.5526 % | 75.5499 % | 57 |

There is a difference in the content since the Logics Regression is used for classification and Elastic Net is used for prediction or regression. Fig. 8 shows the ROC curve with AUC =0.801 produced by Logistic Regression using the Weka tool.

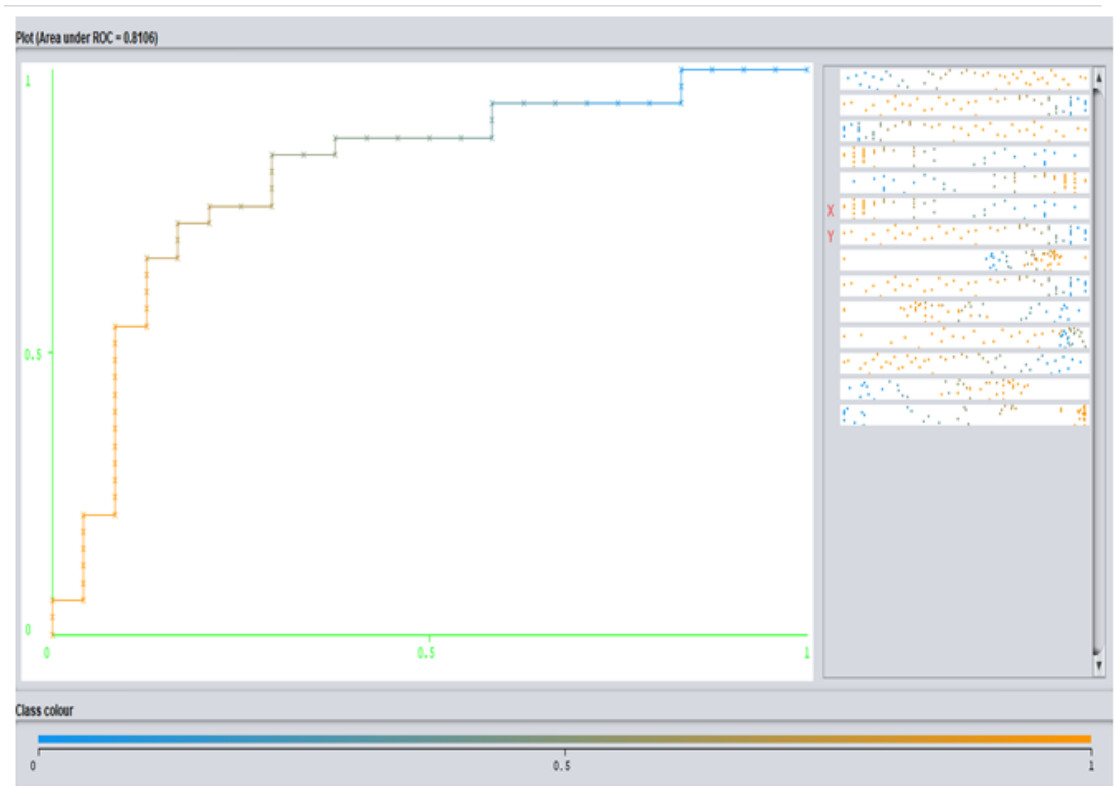

**Fig. 8. Shows ROC curve produced by the Logistic Regression model.**

This value for the Logistic Regression model is good if it is compared to the M-RNNGWO ROC curve with AUC=0.872, which is close to excellent. In other words, the M-RNNGWO model is steadier and its AUC is far from the baseline, which represents the ROC curve of a random predictor: it has an ROC with AUC of 0.5. Thus, this proves that our models are useful [33,34].

## Conclusion

In this paper, a student performance system was suggested for classifying students in English courses based on their previous accomplishments, social setting, and academic setting. The classification technique used a modified GWO for optimization of weights and biases of a modified RNN model. The modification in the GWO involved inserting another best solution into the population of the wolves. Also, the average of the distance of the best wolves was taken into consideration instead of taking the separate distances of the best wolves. This modification had a good effect, since the position of the search agents was updated with an extra best solution. The concept involved the simple RNN type based on an MLP with two hidden layers as a classifier for the prediction of student outcomes. In general, the aim of using meta-heuristic methods with a neural network is to maximize the outcome of the neural network model. The results demonstrated that the proposed adaptation enhanced the students' performance positively.

Depending on the obtained results, the M-RNNGWO is compared with several proposed models not limited to the Logistic Regression and Elastic Network, and the M-RNNGWO produced an accuracy of 98.6% and outperformed some other algorithms. Also, M-RNNGWO produced a ROC AUC of 0.872, which means that the model is close to perfect; it also indicates that the classification results of the M-RNNGWO are statistically significant. This level of accuracy indicates that the M-RNNGWO was found to be more stable in terms of encountering the overfitting problem and handling the local minima problem.

## Acknowledgment

The authors would like to thank the journal's editorial office and distinct reviewers for their effort and time for revising the paper. The authors would like to thank the American Journal Experts (AJE) for editing this paper. Special thanks are sent to both Dr. Edward Bassett from the English Language Centre (a Juris Doctorate from the University of Missouri-Columbia Law School (USA) and a Master's in Fine Arts (Creative Writing) from the University of Southern Maine (USA) and Mr. Shalaw Najat Ghani (MA in TESOL), from Valparaiso University, for their continuous effort in editing the manuscript.